\documentclass[10pt,twocolumn,letterpaper]{article}
\pdfoutput=1
\usepackage[pagenumbers]{cvpr} 

\usepackage{graphicx}
\usepackage{amsmath}
\usepackage{amssymb}
\usepackage{booktabs}
\usepackage[accsupp]{axessibility}
\usepackage[pagebackref,breaklinks,colorlinks]{hyperref}

\usepackage[capitalize]{cleveref}
\crefname{section}{Sec.}{Secs.}
\Crefname{section}{Section}{Sections}
\Crefname{table}{Table}{Tables}
\crefname{table}{Tab.}{Tabs.}

\def\cvprPaperID{8296} 
\def\confName{CVPR}
\def\confYear{2022}

\newcommand{\margin}{\vspace{3pt}\noindent}

\begin{document}

\title{Represent, Compare, and Learn: A Similarity-Aware Framework for\\Class-Agnostic Counting}

\author{Min Shi ~~~
\quad Hao Lu ~~~
\quad Chen Feng ~~~
\quad Chengxin Liu ~~~ 
\quad Zhiguo Cao\thanks{Corresponding author}\\
School of Artificial Intelligence and Automation, Huazhong University of Science and Technology, China\\
$\tt\footnotesize \{min\_shi, hlu, chen\_feng, cx\_liu, zgcao\}@hust.edu.cn$
}

\maketitle
\begin{abstract}
Class-agnostic counting (CAC) aims to count all 
instances in a query image given 
few exemplars. A standard pipeline is to 
extract visual features from exemplars and 
match them with query images to infer object counts. Two essential components in this pipeline are feature representation and similarity metric. Existing methods either adopt a pretrained network to represent features or learn a new one, while applying a naive similarity metric with fixed inner product. We find this paradigm leads to noisy similarity matching and hence harms counting performance. In this work, we propose a similarity-aware CAC framework that jointly learns representation and similarity metric. 
We first instantiate our framework with a naive baseline called Bilinear Matching Network (BMNet),
whose key component is a learnable bilinear similarity metric. To further embody the core of our framework, we extend BMNet to BMNet+ 
that models similarity from three aspects: 1) \textbf{representing} the instances via their self-similarity to enhance feature robustness against intra-class variations; 2) \textbf{comparing} the similarity dynamically to focus on the key patterns of each exemplar; 3) \textbf{learning} from a supervision signal to impose explicit constraints on matching results.  Extensive experiments on a recent CAC dataset FSC147 show that our models significantly outperform state-of-the-art CAC approaches. In addition, we also validate the cross-dataset
generality of BMNet and BMNet+ on a car counting dataset CARPK. Code 
is at {\tt tiny.one/BMNet} 

\end{abstract}

\section{Introduction}
\label{sec:intro}
Object counting aims to infer the number of objects from an image. Most existing methods focus on a specific category, \eg, crowd~\cite{zhang_2015_CVPR}, animal~\cite{count_animals}, or car~\cite{car_counting}, while requiring numerous training data to learn a good model. In contrast, given only one exemplar of a novel category, \eg, a car, even a child can easily capture its visual properties and count cars in new scenes. Recently, CAC (Class Agnostic Counting)~\cite{gmn, cfocnet, famnet}, which counts objects of arbitrary categories given only 
few exemplars, is proposed to reduce 
the 
reliance on training data. CAC points out a promising direction for object counting, \ie, from learning to count objects to learning the way to count.
\begin{figure}[t]
  \centering
   \includegraphics[width=1.0\linewidth]{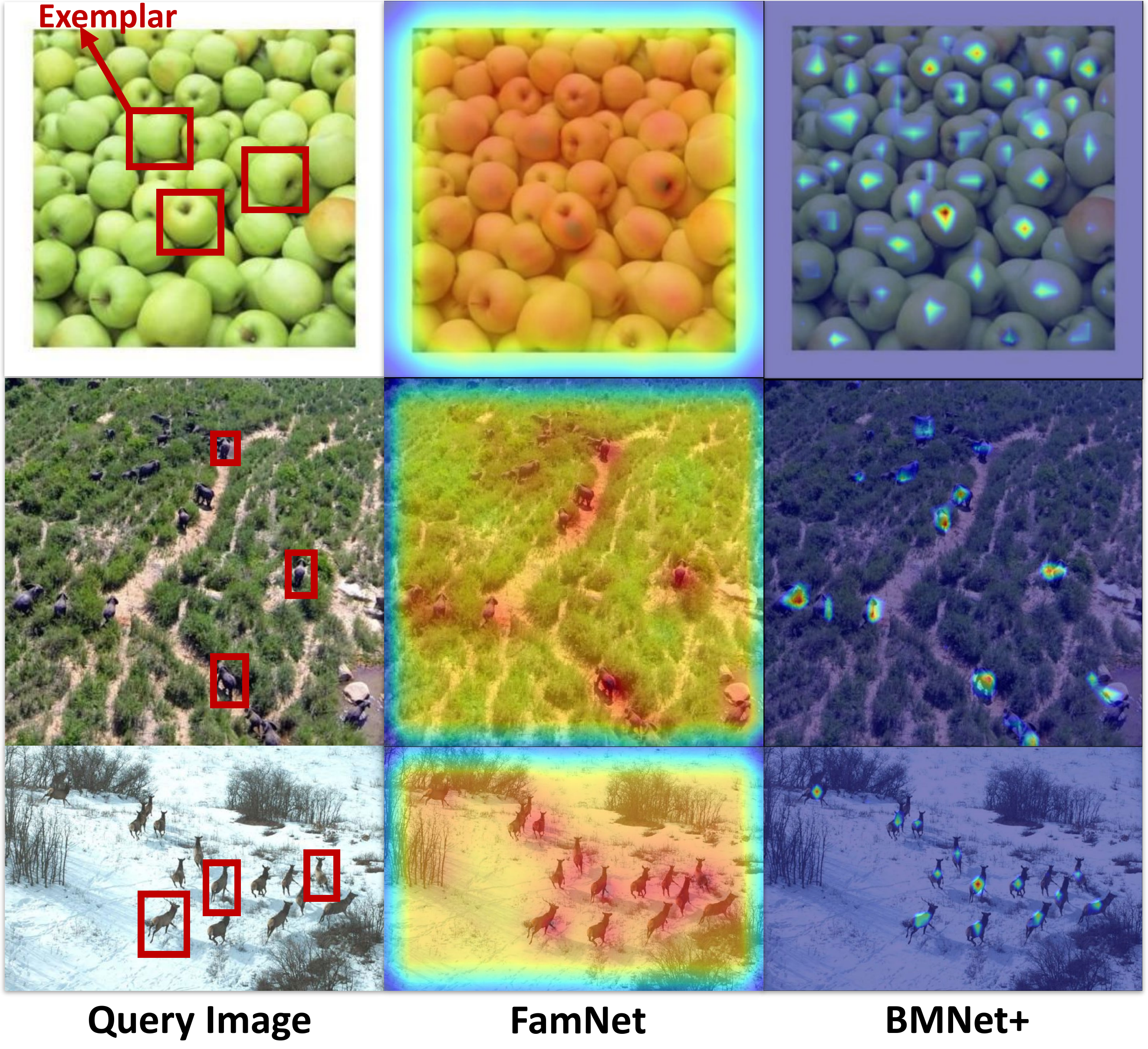}\vspace{-5pt}
   \caption{\textbf{Visualizations of intermediate similarity maps in class-agnostic counting}. Compared with the state-of-the-art FamNet~\cite{famnet}, our model (BMNet+) generates high-fidelity results.}
   \label{fig:corr-map}
   \vspace{-5pt}
\end{figure}

Generally, existing CAC methods~\cite{gmn, cfocnet, famnet} work in an \textit{extract-and-match} pipeline. They first extract visual features from exemplars and match these features with those of query images. Similarity matching results are then used as intermediate representations to infer object counts. Intuitively, two factors play 
critical roles: feature representation and similarity metric. Existing methods either use a learnable~\cite{gmn, cfocnet} or a fixed feature extractor~\cite{famnet}, but 
apply a similarity metric 
with 
some pre-defined rules, \textit{e.g.}, inner product~\cite{cfocnet, famnet}. We find this 
can yield unsatisfactory matching results. From Fig~\ref{fig:corr-map}, by examining a recent model FamNet~\cite{famnet}, we observe obvious noise on 
background 
and 
weak responses on target positions. The resulting density map may be erroneous given such ambiguity. 

In this work, we present a generic similarity-aware framework for CAC, which jointly learns representation and similarity metric in an end-to-end manner. Our goal is to seek better similarity modeling that can generalize well to novel categories. First, 
we instantiate a bilinear matching network (BMNet), which extends the fixed inner product to a learnable bilinear similarity metric and also allows 
learnable representation 
through back-propagation. Unlike fixed inner product, the bilinear similarity metric captures flexible interactions among feature channels to measure similarity. Then, we extend BMNet to BMNet+ to embody the core motivation of our framework from three aspects: 
\textit{representing} instances via self-similarity, \textit{comparing} the similarity dynamically, and \textit{learning} with explicit, similarity-aware supervision. 
In particular, we apply self-attention~\cite{sa_gan} to represent self-similarity among features to mitigate intra-class variations. It augments the feature of each instance with information from other intra-class instances 
such that 
complementary clues like scales or viewpoints can be offered.
The dynamic similarity metric applies a feature selection module to the exemplars to find key patterns and hence embraces both dynamism and selectivity. Then, inspired by metric learning~\cite{metric_learning_reality_check}, the similarity loss imposes an explicit supervision on the intermediate similarity map to pull the exemplar and the target close but 
to 
push the exemplar and background away. 

Experiments on the public benchmark FSC147~\cite{famnet} show that our method outperforms the previous best 
approaches by large margins, with a relative improvement of $+33.72\%$ and $+33.79\%$ on the validation and test sets in terms of mean absolute error. According to Fig.~\ref{fig:corr-map}, our method outputs better intermediate similarity results and presents generality over different categories. The ablation study validates the three main components within BMNet+. And we further show the cross-dataset generality of our models on a car counting dataset CARPK~\cite{carpk}.

Our contributions are two-fold:

$\bullet$ A generic CAC framework that includes the existing pipeline and also generalizes it with joint representation learning and similarity 
learning;

$\bullet$ BMNet and BMNet+: two CAC models instantiated from our framework, which models packed similarity.

\section{Related Work}
\begin{figure*}[!t]
  \centering
   \includegraphics[width=1.0\linewidth]{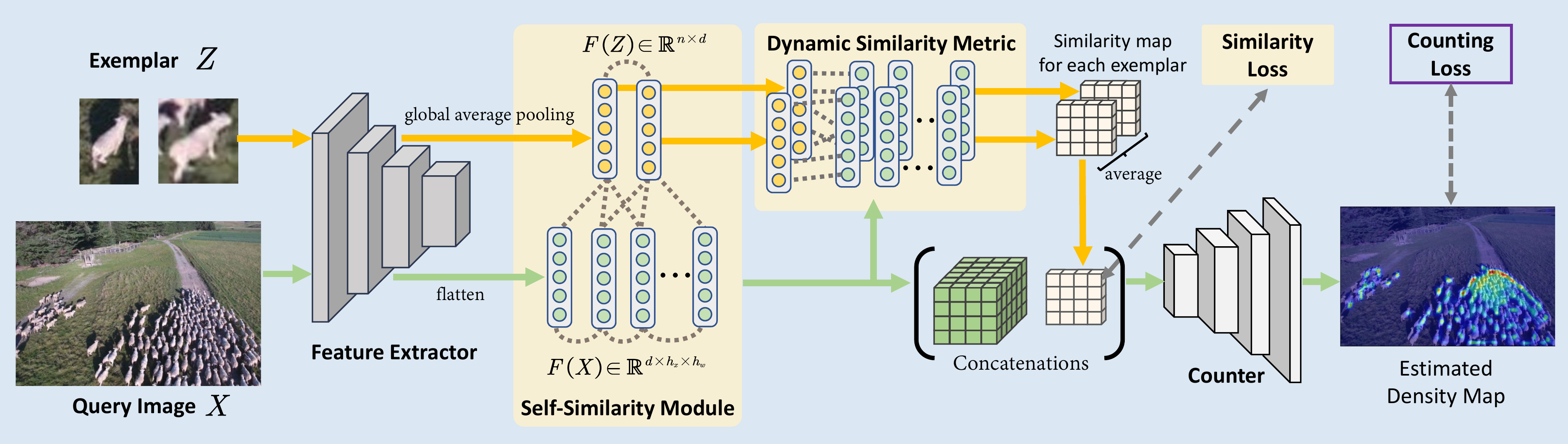}\vspace{-5pt}
   \caption{\textbf{The pipeline of BMNet and BMNet+}. BMNet follows the \textit{extract-then-match} paradigm but learns representation and similarity metric jointly in an end-to-end manner. BMNet+ is an improved version whose differences from BMNet are highlighted in colored 
   blocks.}
   \label{fig:baseline-pipeline}
   \vspace{-5pt}
\end{figure*}

\subsection{Class-Specific Object Counting}
According to how the counting problem is formulated, existing methods can be categorized into counting by detection~\cite{counting_by_detection}, regression~\cite{count_by_regression, zhang_2015_CVPR, zhang2016singleMCNN, dmcount}, classification~\cite{bcnet}, and localization~\cite{d2cnet, p2pnet, topocount}.
The most-studied regression-based approaches formulate counting as a dense prediction~\cite{indexnet, indexnet_pami} task, which learns to predict density maps~\cite{density_map}. Under this paradigm, most methods focus on designing network architectures~\cite{zhang2016singleMCNN}, multi-scale strategies~\cite{sdcnet, sasnet}, or new loss functions and learning targets~\cite{dmcount, bloss}. Recently, new paradigms are developed such as reinforcement learning~\cite{libranet} and counting by localization~\cite{topocount, d2cnet, p2pnet}. The key difference between class-specific counting and class-agnostic one lies in that the latter requires a more generic representation and a more discriminative similarity metric. 

\subsection{Class-Agnostic Counting}
Lu \textit{et al.}~\cite{gmn} first address CAC and propose a general matching network. One convolutional neural network is shared to extract feature maps for both query images and exemplars. These features are then concatenated to regress the object count. Considering that direct regression from concatenated features may cause overfitting, recent methods start to model similarity explicitly. CFOCNet~\cite{cfocnet} uses the feature map of exemplar as a 2D kernel to convolve over the query feature map, following the spirit of Siamese network in object tracking~\cite{siamesenet}. They also design a multi-scale matching framework to improve robustness. FamNet~\cite{famnet} also adopts siamese way to model similarity and further proposes test-time adaptation given test exemplars. To alleviate the shortage of training data, Ranjan \textit{et al.}~\cite{famnet} propose the first and only CAC dataset FSC147 that covers challenges like occlusion and scale variation. The above methods report promising results for CAC. However, they typically focus on multi-scale strategy, data amplification, or test-time adaptation, but neglect a fundamental problem -- similarity modeling. In this work, we show the importance of similarity modeling and also present a generic framework that jointly learns both representation and similarity metric. 

\subsection{Metric Learning}
Metric learning aims to embed data into a space where similar samples are pulled close and dissimilar ones are pushed away~\cite{metric_learning_reality_check}. The similarity is measured in a fixed ~\cite{NCC1,LPNorm1} or learned~\cite{cosface, circle_loss} manner. One common way constrains the similarity between features in a pair~\cite{contrastive_loss} or triplet~\cite{ triplet_loss}. Another way adds constraints based on signal-to-noise ratios~\cite{snr_loss, info_nce, np_loss}, where the similarity between positive sample pairs is considered as the signal, and the similarity between negative pairs as the noise. The idea is to strengthen the signal and weaken the noise. We repurpose this idea into CAC, \ie, pulling close the features between the exemplar and target instances, while pushing away the features between exemplars and background patches. We further design a similarity loss based on this idea to supervise the similarity matching results. 

\section{A Similarity-Aware Framework for Class-Agnostic Counting}
\begin{figure}[t]
  \centering
   \includegraphics[width=1.0\linewidth]{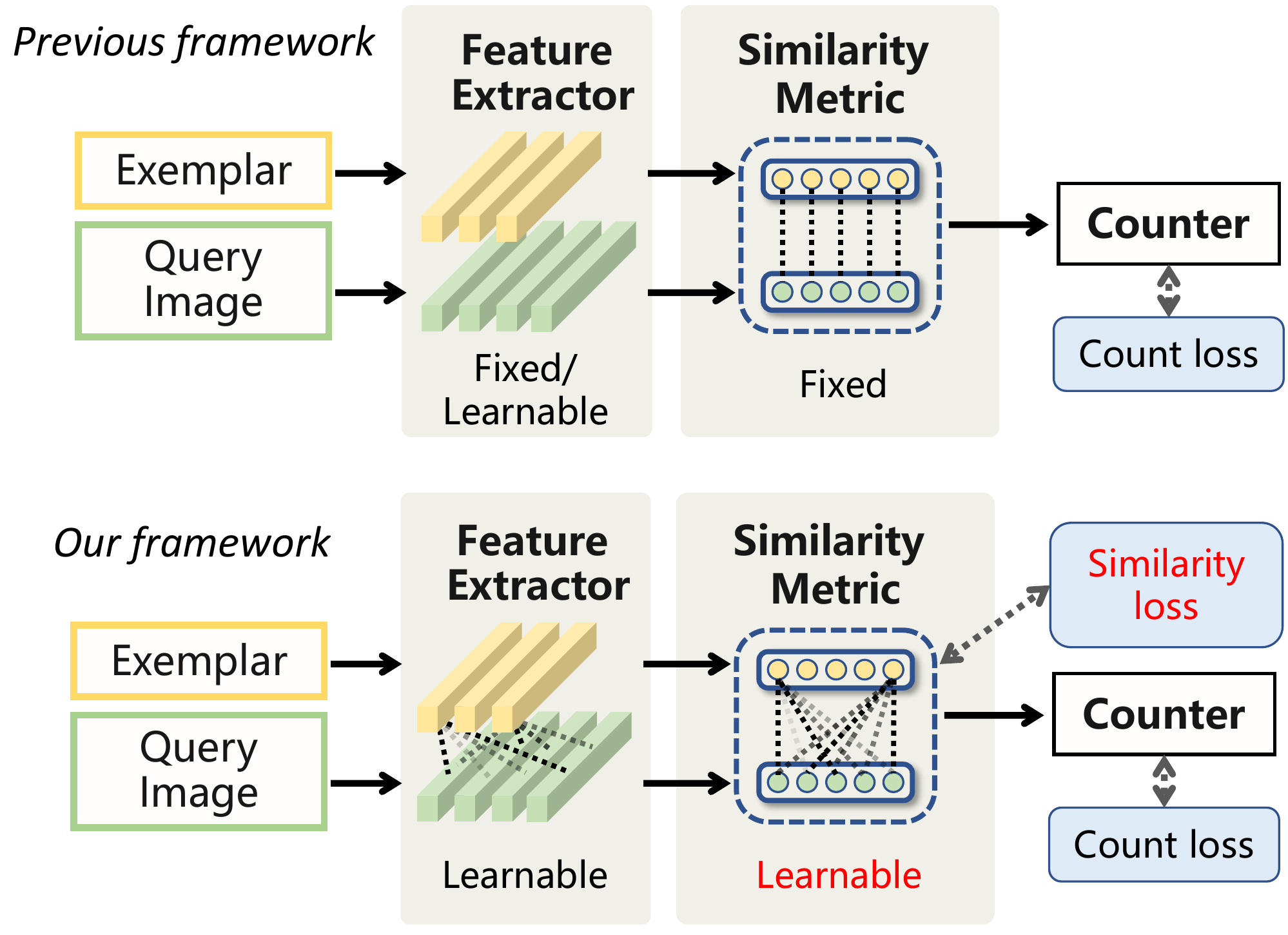}\vspace{-5pt}
   \caption{\textbf{The comparison between the previous framework and ours}. Ours can learn both representation and similarity metric jointly with more flexibility and generality.}
   \label{fig:basic-framework}
   \vspace{-5pt}
\end{figure}

This section presents our framework for CAC, which jointly learns representation and similarity metric in an end-to-end manner (Fig.~\ref{fig:basic-framework}). We first instantiate this framework with a naive baseline, termed Bilinear Matching Network (BMNet), and then propose an extended BMNet+ to exemplify our idea on how to represent, dynamize, and learn similarity for both representation and similarity metric. The detailed pipeline of our methods is in Fig.~\ref{fig:baseline-pipeline}.

\subsection{Bilinear Matching Network}
\label{sec:bmn}
Differing from previous CAC methods, BMNet allows simultaneous optimization of representation and similarity metric. The core of BMNet is the bilinear similarity metric that captures flexible interactions among feature channels to model similarity. 

Given a query image $X$ and an exemplar $Z$ of arbitrary category $c$, CAC aims to count all the instances of category $c$ within $X$. Without loss of generality, we use one exemplar to explain our pipeline (we will also note how to operate with multiple exemplars). 

\margin
\textbf{Feature Extractor. }
The feature extractor consists of layers of convolutional operations that map the input into $d$-channel features. For the query $X$, it outputs a downsampled feature map $F(X)\in \mathbb{R}^{d\times h_x\times w_x}$. For the exemplar $Z$, the output feature map is further processed with global average pooling to form a feature vector $F(Z)\in \mathbb{R}^{d}$. 

\margin
\textbf{Learning Bilinear Similarity Metric.}
Previous methods apply fixed inner product to compute the similarity between two feature vectors. We argue that such fixed one-to-one interactions may be insufficient in modeling class-agnostic similarity. Inspired by neural similarity learning~\cite{neural_similarity_learning} and bilinear models~\cite{low_rank_bp}, we propose to extend the original inner product to a learnable bilinear similarity, which establishes flexible connections between two vectors. Specifically, let $F_{ij}(X)\in\mathbb{R}^d$ be the channel feature at spatial position $(i,j)$. By redefining $\boldsymbol{x}_{ij}=F_{ij}(X)$ and $\boldsymbol{z}=F(Z)$, the similarity map $S$ can be obtained by
\begin{equation}
\label{equation: learnable-innerproduct}
S_{ij}\left(\boldsymbol{x,z}\right) = \left( P\boldsymbol{x}_{ij} + \boldsymbol{b}_x \right) ^T\left( Q\boldsymbol{z} + \boldsymbol{b}_z \right)\,, 
\end{equation}
where $P,Q \in \mathbb{R}^{d\times d}$ are learnable matrices, and $\boldsymbol{b}_x,\boldsymbol{b}_z \in \mathbb{R}^{d\times 1}$ are learnable biases. The initial bilinear metric is in the form $\boldsymbol{x}^{T}W\boldsymbol{z}$. We decompose $W$ into $P,Q$ specific to the query image and the exemplar, respectively. In practice, we find that this can yield better performance (refer to supplementary material for more details). 

Given $n$ exemplars, one can use Eq.~\ref{equation: learnable-innerproduct} repetitively to compute $n$ similarity maps, and then output their averaged similarity as the final similarity map $S$.

\margin
\textbf{Counter.}
The counter receives the channel-wise concatenation of the query feature map $F(X)$ and the similarity map $S$, and then predicts a density map $\boldsymbol{D}_{\rm pr}$. The final count is the integral of $\boldsymbol{D}_{\rm pr}$. In practice, the counter consists of convolutional and bilinear upsampling layers.

\margin
\textbf{Supervision Signal.}
We adopt a conventional $\ell_2$ loss as the counting loss $\mathcal{L}_{\rm count}$:
\begin{equation}
\mathcal{L}_{\rm count} = ||\boldsymbol{D}_{\rm pr}(X,Z)-\boldsymbol{D}_{\rm gt}(X,Z)||_2^2\,,
\end{equation}
where $\boldsymbol{D}_{\rm gt}$ denotes the ground truth density map. 

\subsection{Learning Dynamic Similarity Metric}
\label{sec:dynamic-similarity-metric}
\begin{figure}[!t]
  \centering 
   \includegraphics[width=1\linewidth]{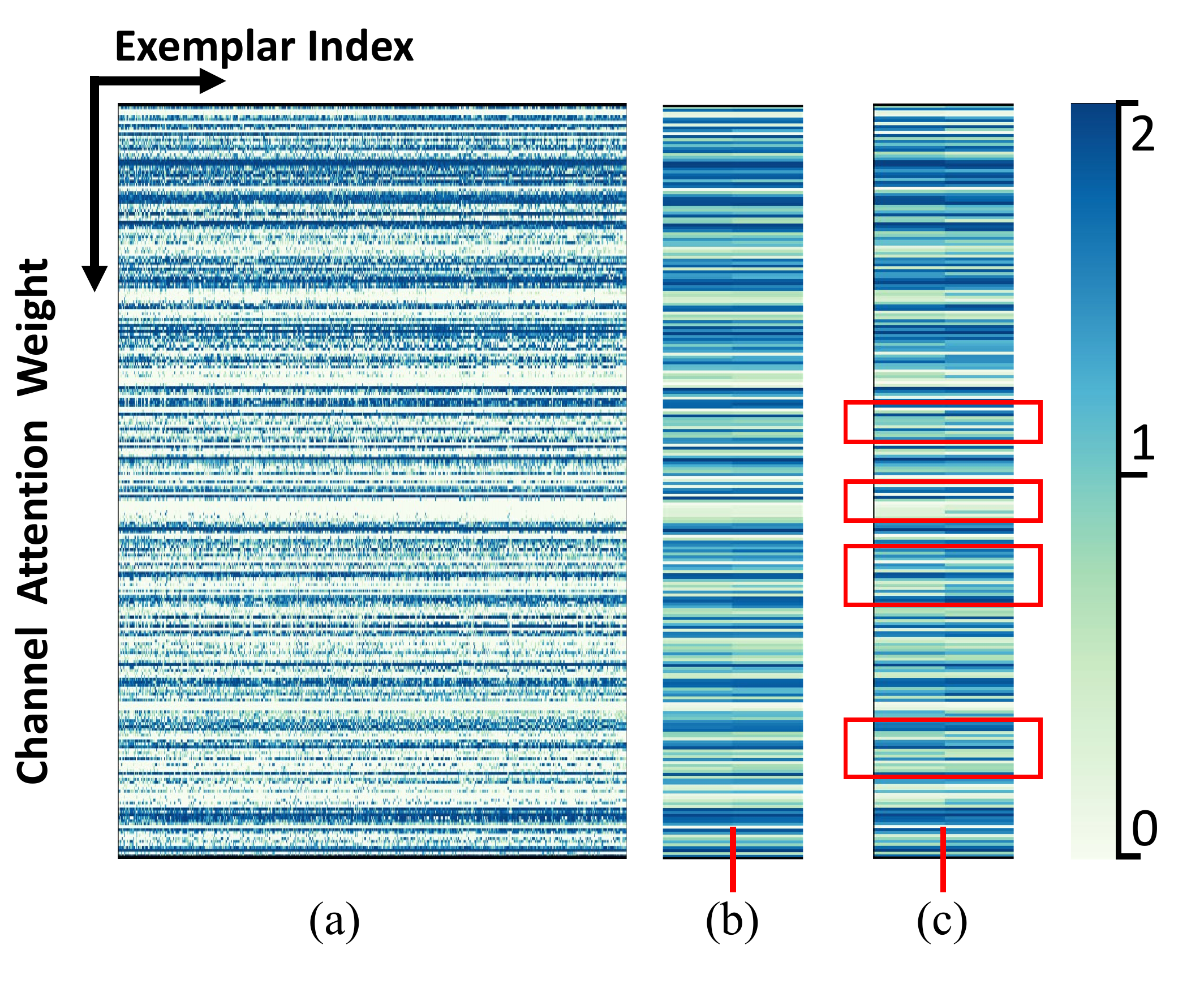}\vspace{-10pt}
   \caption{\textbf{Visualization of channel attention weights for exemplars from the same and different categories.}
   We visualize the attention weights (each vertical line) for exemplars from (a) the same category \textit{apple}, (b) visually similar categories \textit{apple vs. strawberry}, and (c) dissimilar ones \textit{apple vs. stamp}. For (b) and (c), the red short line splits the samples into two categories. By focusing on the horizontal lines, we can observe that, channel attention weights for exemplars of the same or similar categories (cf. (a) and (b)) shows more consistency than those for different categories (cf. (c), especially in red boxes). Better zoom in for details. More visualizations can be found in supplementary material.}
   \label{fig:dynamic-weight-visualization}
   \vspace{-5pt}
\end{figure}

The bilinear similarity in Sec.~\ref{sec:bmn} increases flexibility to model similarity. However, the learned similarity metric stays fixed once trained and treats all categories equally during inference. Considering that humans may learn to recognize a category based on category-specific patterns, \eg, if told something is furry with four legs and pointy ears, one may suppose it to be a cat. We therefore think it is better to develop a dynamic similarity metric that can adaptively learn to focus on the key patterns of exemplars. Inspired by this intuition, we integrate a feature selection module over the exemplars to generate an exemplar-specific metric. 
Specifically, we regard each channel in $Q\boldsymbol{z} + \boldsymbol{b}_z$ as a pattern. Similar to SENet~\cite{senet}, we learn the dynamic channel attention weight $\boldsymbol{a}$ conditioned on $Q\boldsymbol{z} + \boldsymbol{b}_z$ such that similarity $S$ can be computed by
\begin{equation}\label{equation:dynamic-similarity-measuring-metric}
    S_{ij}\left(\boldsymbol{x,z}\right) = \left[ (P\boldsymbol{x}_{ij} + \boldsymbol{b}_x) \right] ^T\left[\boldsymbol{a}\circ (Q\boldsymbol{z} + \boldsymbol{b}_x) \right]\,, 
\end{equation}
where $\circ$ denotes the Hadamard product. 

We exemplify the learned dynamic attention weights in Fig.~\ref{fig:dynamic-weight-visualization}. For exemplars of the same category (cf. Fig.~\ref{fig:dynamic-weight-visualization}(a)), the generated dynamic attention weights turn similar. Similar phenomena can be observed given two visually close categories (cf. Fig.~\ref{fig:dynamic-weight-visualization}(b)). This validates our intuition that the dynamic similarity metric learns to focus on similar visual patterns for similar categories. In contrast, given two visually different categories (cf. Fig.~\ref{fig:dynamic-weight-visualization}(c)), our method learns to extract different key patterns with clear distinction. Note that whatever the cases are, there exist common patterns between different categories. This accords with the way we humans recognize objects: first use general visual clues like shapes and colors, then focus on category-specific details. 

\subsection{Supervising the Similarity Map}

\begin{figure}[!t]
  \centering
   \includegraphics[width=1\linewidth]{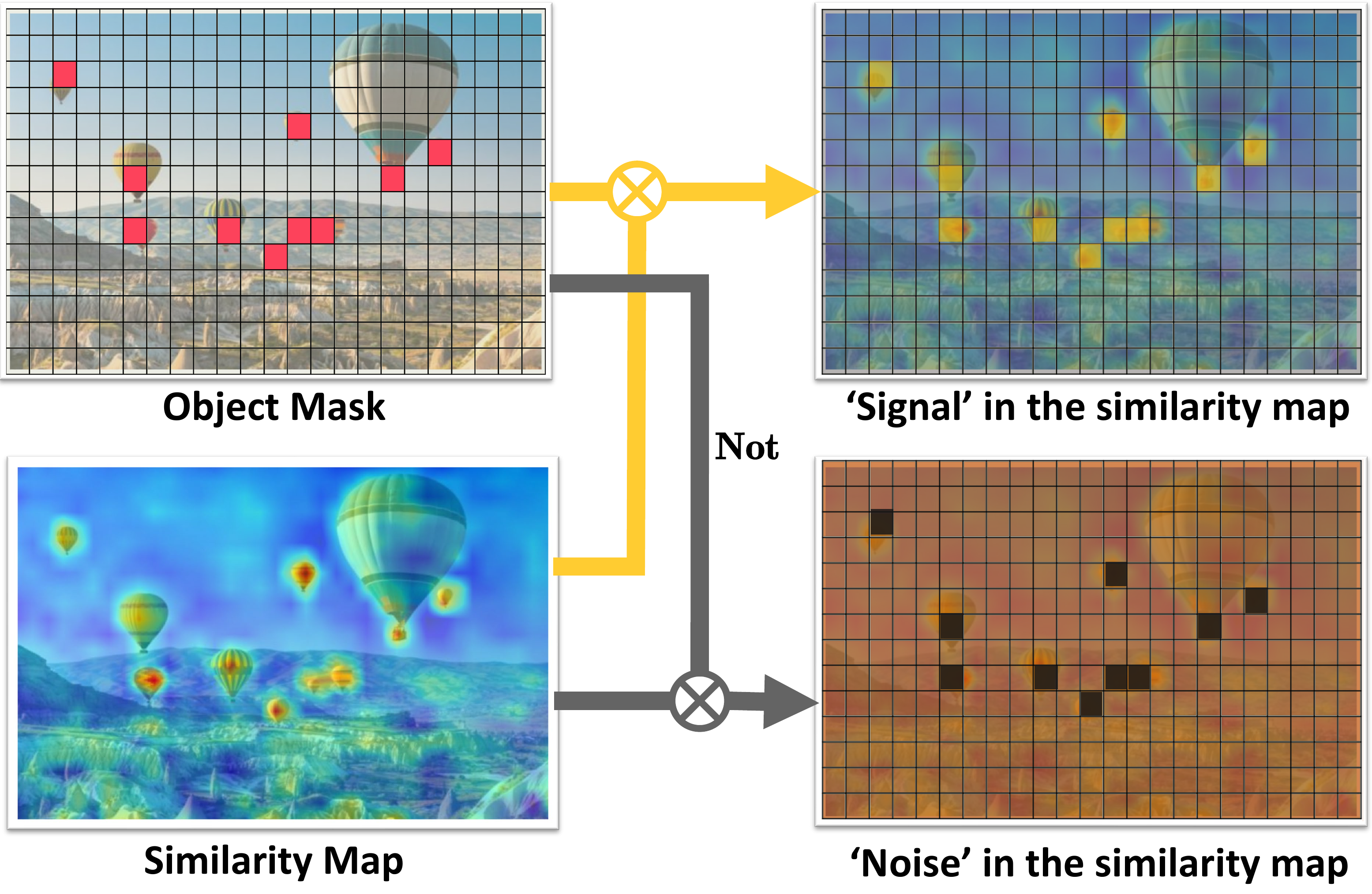}\vspace{-5pt}
   \caption{\textbf{An 
   illustration on how to compute the similarity loss.} We consider the similarity between exemplars and target instances as signals (positive labels), while similarity between exemplars and background as noises (negative labels).}
   \label{fig:feature-generation-loss}
   \vspace{-5pt}
\end{figure}

Both existing CAC methods and our baseline BMNet only use the counting loss as supervision during training. In practice, we find that direct supervision on similarity matching results can help to guide similarity modeling. To this end, we start by posing a fundamental question: \textit{what makes an ideal similarity metric for CAC?} In our opinion, it should output high similarity between the two features of the same category and low one for differing categories. This accords with the idea of metric learning~\cite{metric_learning_reality_check}.

Here we present a simple way to achieve this. Suppose the size of $S$ is $1/r$ of that of $X$, \ie, each position in similarity map corresponds to a $r\times r$ block within query image. For each position in $S$, we assign a positive label if its corresponding $r\times r$ block contains more than one target, and assign a negative one if it contains no target. We then derive the similarity loss $\mathcal{L}_{\rm sim}$ with signal-to-noise ratio:
\begin{equation}
\mathcal{L}_{\rm sim} =-\log \frac{\sum_{i\in\rm pos}{\exp \left(S_i\right)}}{\sum_{i\in\rm pos}{\exp \left( S_i \right)}+\sum_{j\in\rm neg}{\exp \left(S_j \right)}}\,.
\end{equation}
Here $pos, neg$ denotes positive and negative positions in $S$.

With the counting loss $\mathcal{L}_{\rm count}$ and the similarity loss $\mathcal{L}_{\rm sim}$, the final training loss can be written as 
\begin{equation}
\label{eq:total-loss}
    \mathcal{L} = \mathcal{L}_{\rm count}(D_{\rm pr}, D_{\rm gt}) + \alpha \cdot \mathcal{L}_{\rm sim}(S)\,,
\end{equation}
where $\alpha$ balances the two component loss items.

\subsection{Self-Similarity Module}
\begin{figure}[t]
    \centering
    \includegraphics[width=1.0\linewidth]{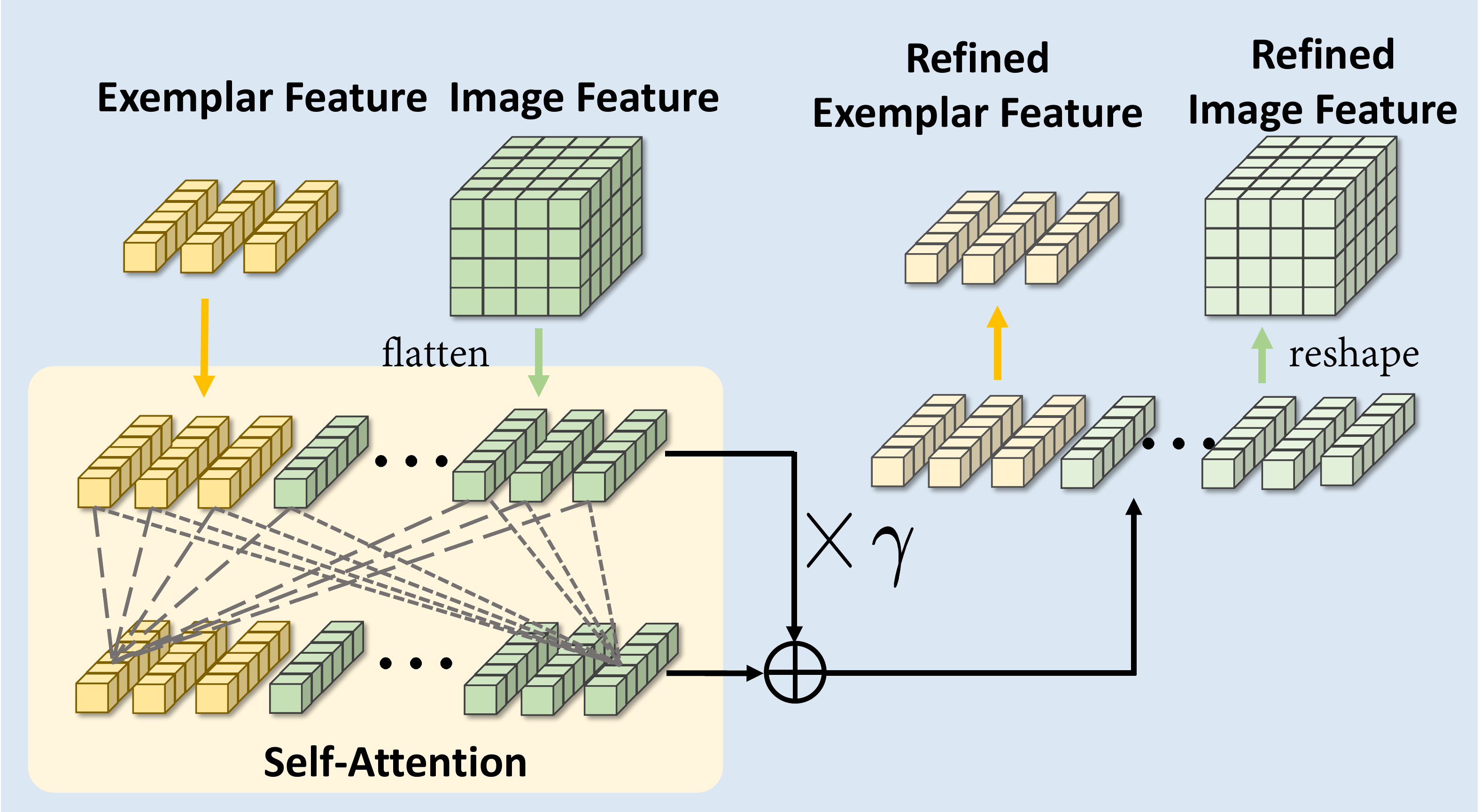}
    \vspace{-15pt}
    \caption{\textbf{Self-similarity module.}}
    \label{fig:self-attention-layer}
    \vspace{-5pt}
\end{figure}

\begin{figure}[t]
  \centering
   \includegraphics[width=1\linewidth]{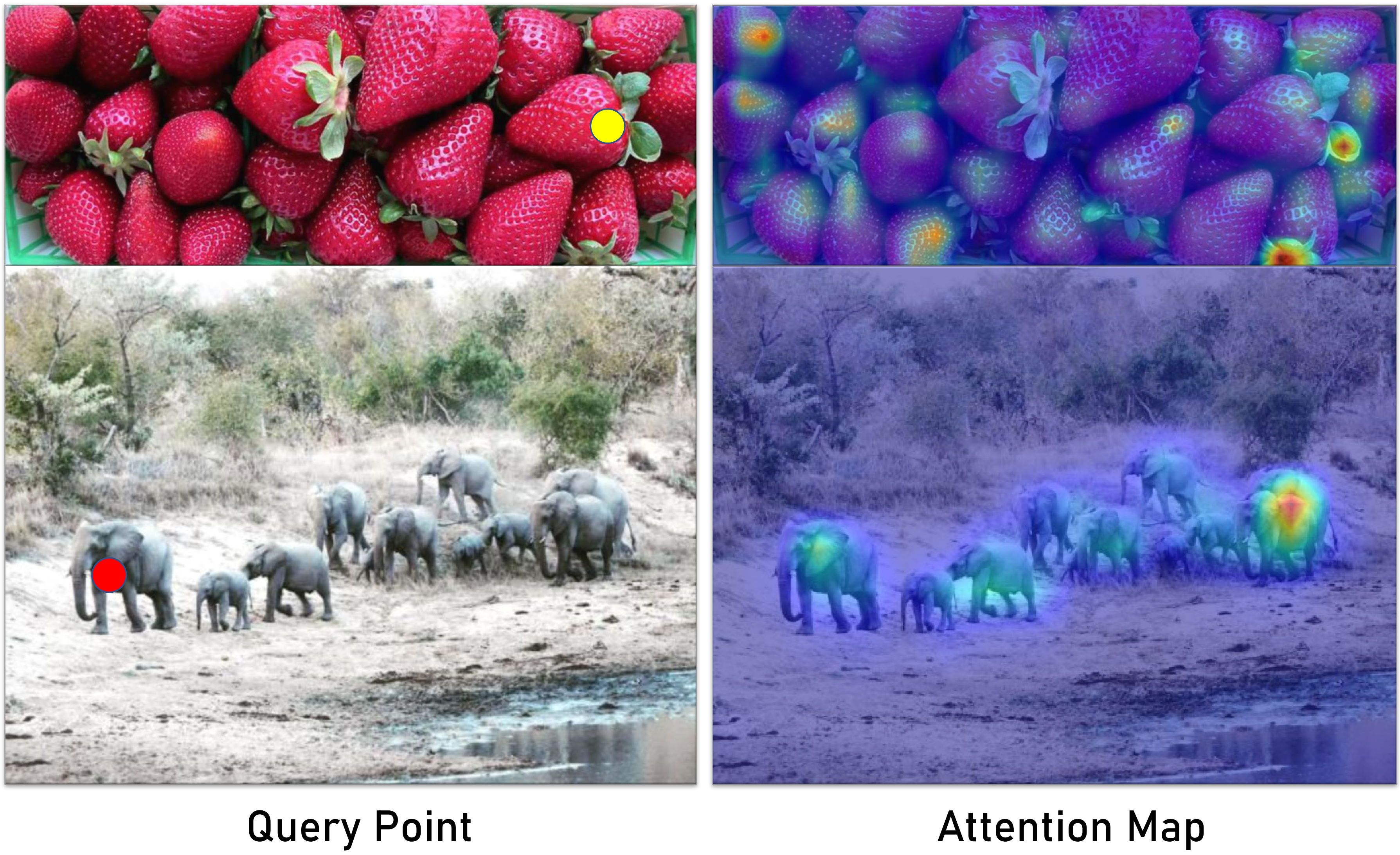}\vspace{-5pt}
   \caption{\textbf{Exemplified attention maps using self-similarity map.}}
   \label{fig:self-attention-visualization}
   \vspace{-5pt}
\end{figure}

The core of our framework also includes improving the representation suitable for similarity matching. Here we present a feasible way to address this. 
As in Fig.~\ref{fig:self-attention-visualization}, in reality, instances of the same category often appear with different attributes like poses and scales. Such intra-class variations impose great challenges on similarity matching. Accordingly, we propose to augment each instance feature with complementary information from other instances of the same category but with different attributes. 

Technically, we first collect the exemplar feature $F(Z)$ and each feature vector $F_{ij}(X)$ from the query feature map into a feature set. Then each vector in the feature set is updated via a self-attention mechanism~\cite{sa_gan} (Fig.~\ref{fig:self-attention-layer}). 
The updated features are added back to the original ones with a learnable ratio $\gamma$. The resulting feature set is then re-split and re-shaped to obtain the final $F(Z)$ and $F(X)$.

We remark that, \cite{cfocnet} also applies self-attention over the feature maps similar to our work; hence the self-similarity module does not constitute our contribution. However, here we attempt to explain how self-attention works in our task. We start by visualizing the self-attention maps given the query points as in Fig.~\ref{fig:self-attention-visualization}. It can be observed that each query point mainly focuses on instances of the same category. This differs from self-attention in object detection~\cite{detr} where the query point mainly focuses on a single instance. This indicates that, the self-similarity module in CAC tends to aggregate same-category information and hence enhances representations with robustness towards intra-class variations.

\margin
\textbf{Scale Embedding}. 
Inspired by the positional embedding in Transformer~\cite{transformer}, we wonder if we could similarly embed the scale information of the exemplars to improve the representation. Note that two factors cause the exemplars to lose scale information in our method: one is the resizing of exemplars and the other is the pooling operation during feature extracting. 
To compensate for this loss, we propose to augment the exemplar's feature with its corresponding scale embedding. We discretize the scale space into $l_{\rm total}$ levels. Each scale level is assigned with a $d$-dimensional embedding vector, yielding an embedding set whose cardinality equals $l_{\rm total}$. Given an exemplar $Z$ and query image $X$, we first derive $Z$'s scale level $l(Z)$ by
\begin{equation}\label{equation:level}
    l(Z)=\min \left( l_{\rm total}-1, \lfloor \left( \frac{h_Z}{2h_X}+\frac{w_Z}{2w_X} \right) \cdot l_{\rm total} \rfloor \right)\,,
\end{equation}
where $h_Z, w_Z, h_X, w_X$ denote images' heights and widths. Then the scale embedding vector of level $l(Z)$ is retrieved and added back into the original feature. The scale embedding set is randomly initialized and learned during training, and stays fixed during inference. 
\subsection{Implementation Details}
For a fair comparison, we apply the same pre-processing to query images and the feature extractor as in FamNet~\cite{famnet}.

\margin
\textbf{Data Pre-processing.}
We resize the query image while keeping its aspect ratio so that the length of its sides is limited within $[384, 1584]$. Exemplars are resized to $128\times 128$ before fed into the feature extractor. No data augmentation is applied. During training, the size of all query images within a mini-batch is kept the same by zero-padding. 

\margin
\textbf{Network Architecture.}
The feature backbone consists of the first $4$ blocks of ResNet-50~\cite{resnet}, which outputs the feature maps of $1024$ channels. For each query image, the number of channels are reduced to $256$ using $1 \times 1$ convolution. For each exemplar, the feature maps are first processed with global average pooling and then linearly mapped to obtain a $256$D feature vector. The counter consists of a few convolution and bilinear upsampling layers to regress a density map of the same size as the query image. When computing channel attention weight $\boldsymbol{a}$ in BMNet+, we apply a Linear(128)-ReLU-Linear(256)-Tanh structure, where the number in the bracket denotes the output dimension. Refer to supplementary material for more details.

\margin
\textbf{Training Details.}
Our model is trained end-to-end. The backbone is initialized via SwAV~\cite{swav}. Other parameters are randomly initialized. We apply AdamW~\cite{adamw} as the optimizer with a batch size of $8$. The model is trained for $300$ epochs with a fixed learning rate of $1e$-$5$. The weight $\alpha$ of similarity loss in Eq.~\ref{eq:total-loss} is set to $5e$-$6$ so that all the loss items are of the same order of magnitude. The total number of scale levels in Eq.~\ref{equation:level} is empirically set to $20$. We use \texttt{PyTorch}~\cite{paszke2019pytorch} as our experimental platform. Note that the BMNet+ consumes less then $12$GB memory on a single GPU during training. 

\section{Experiments}
\label{exp:all}
Here we first showcase the advantage of our models over the state-of-the-art methods. We then validate each component in BMNet+. Next, we analyze the influence of exemplar numbers and discuss how to integrate features before feeding them to the counter. Finally, we show the cross-dataset generality of our method on a car counting dataset.

\begin{table}
	\centering
	\renewcommand{\arraystretch}{1.2} 
	\addtolength{\tabcolsep}{-4pt}
	\begin{tabular}{@{}lccccc@{}}
		\toprule
		Methods   & Val MAE & Val MSE & Test MAE & Test MSE \\ 
		\midrule
		GMN~\cite{gmn}       & 29.66   & 89.81   & 26.52 & 124.57      \\
		FamNet~\cite{famnet}    & 24.32   & 70.94   & 22.56  & 101.54  \\
		FamNet+~\cite{famnet}   & 23.75   & 69.07   & 22.08  & 99.54   \\
		CFOCNet*~\cite{cfocnet} & 21.19   & 61.41   & 22.10  & 112.71  \\
		BMNet (Ours) & 19.06 & 67.95 & 16.71 & 103.31 \\
		BMNet+ (Ours) & \textbf{15.74} & \textbf{58.53} & \textbf{14.62} & \textbf{91.83} \\ 
		\bottomrule
	\end{tabular}
	\vspace{-10pt}
	\caption{\textbf{Comparison with state of the art on the FSC147 dataset.} Best performance is in boldface.}
	\label{tab:comparison-CAC-methods}
	\vspace{-5pt}
\end{table}

\subsection{Comparison With State of the Arts}
\label{section:comparison-CAC-methods}

\noindent\textbf{The FSC147 Dataset.}
FSC147~\cite{famnet} is the first large-scale dataset for class-agnostic counting. It includes $6,135$ images from $147$ categories varying from animals, kitchen utensils, to vehicles. Given one query image, three instances of the same category are randomly chosen as the exemplars. To validate methods' generality, the categories in training, validation, and test sets have no overlap. All experiments are done on FSC147 unless otherwise specified.

\margin
\textbf{Comparing Methods.}
We mainly compare our models with two available CAC methods: GMN~(General Matching Network~\cite{gmn}) and FamNet~(Few-shot adaptation and matching Network~\cite{famnet}). Since FamNet executes fine-tuning during testing, we denote the fine-tuned version by FamNet+. The other compared methods apply no fine-tuning. Regarding our models, we validate two variants: 1) the baseline BMNet and 2) BMNet+ that implements all core components, \ie, self-similarity module, dynamic similarity metric, and direct supervision on similarity map.
For more comparisons, we also test CFOCNet~\cite{cfocnet} that applies self-attention similar to our work. We reproduce CFOCNet as its code is unavailable and keep the same exemplar pre-processing and training configuration as in our methods. We denote this by CFOCNet*.
Note that the main comparisons are concentrated on the public state-of-the-art FamNet.


\margin
\textbf{Quantitative Results.}
As shown in Table~\ref{tab:comparison-CAC-methods}, BMNet exhibits advantage over all of the compared methods with fixed similarity metrics (FamNet, GMN, and CFOCNet).
Compared with FamNet, BMNet achieves a relative improvement of $21.63\%$ w.r.t.\ validation MAE and $25.93\%$ w.r.t.\ test MAE. Note that BMNet is already a strong baseline over FamNet, which indicates that BMNet is capable of characterizing a novel category without any concerned prior information. One can also observe that BMNet+ reduces the validation MAE by $18.23\%$ and the test MAE by $12.51\%$ compared with BMNet, which validates the effectiveness of our proposed components.

\begin{figure*}[t]
	\centering
	\includegraphics[width=1.0\linewidth]{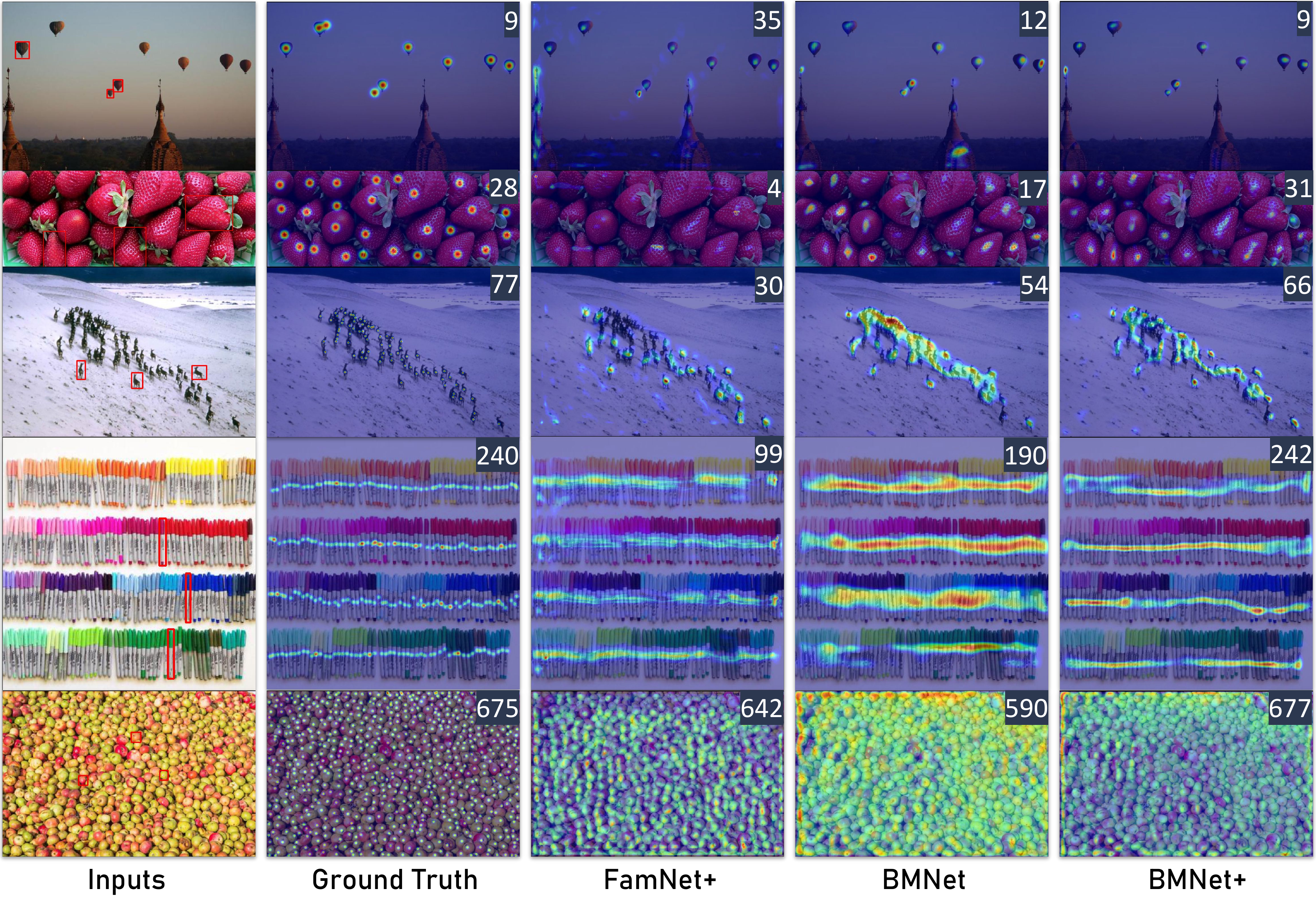}\vspace{-5pt}
	\caption{\textbf{Qualitative results on the FSC147 dataset.} The samples on the left exhibit 
	significant intra-class variations such as scale, pose, and illumination condition. The red boxes 
	indicate exemplars.
	Counting values are shown at the top-right corner. 
	Our BMNet and BMNet+ can predict accurate density maps in both dense and sparse scenes.
	Best viewed by zooming in.
	}
	\label{fig:CAC-visualizations}
	\vspace{-5pt}
\end{figure*}

\margin
\textbf{Qualitative Analysis.}
As shown in Fig.~\ref{fig:CAC-visualizations}, both BMNet and BMNet+ output accurate density maps in whether dense or sparse scenes. Specifically, when counting hot air balloon (the 1st row), FamNet+ and BMNet mistake the tower for counting target, while BMNet+ offers comparatively better discrimination between target and background. In case of strawberries that exhibit large intra-class variation (the 2nd row), FamNet fails while our methods do not. This validates the effectiveness of our bilinear similarity metric (BMNet) and self-similarity in BMNet+. Refer to supplementary materials for more visualizations.

\subsection{Ablation Study on BMNet+}
\label{exp:sl}
Here we justify the effectiveness of each component in BMNet+.
We start by testing the supervision on similarity map, because it directly affects the learning of self-similarity and dynamic similarity metric.

\margin
\textbf{Supervision of the Similarity Map.}
By comparing B1 and B2 in Table~\ref{tab:comparison-feature-refiner}, we can observe that direct supervision on the similarity map brings a relative improvement of $3.88\%$ and $10.41\%$ on the validation and test set w.r.t.\ MAE, respectively. This indicates that the similarity loss can help to learn a generic similarity metric. 

\margin
\textbf{Self-Similarity With Scale Embedding.}
Comparing B2 with B4 in Table~\ref{tab:comparison-feature-refiner}, we can observe that applying self-similarity module and scale embedding improves the validation MAE by $0.97$. However, the performance on the test set shows a converse phenomenon (cf. B7 \textit{vs.} B9). A plausible explanation is that additional parameters in the feature extractor lead to an over-fitting problem. Note that, the comparisons of B3 \textit{vs.} B4 and B8 \textit{vs.} B9 show that scale embedding generally improves representations.

\margin
\textbf{Dynamic Similarity Metric.}
The inclusion of dynamic similarity metric further brings a relative improvement of $9.28\%$ on validation MAE and $11.29\%$ on test MAE (cf. B4 \textit{vs.} B5 and B9 \textit{vs.} B10). In Sec.~\ref{sec:dynamic-similarity-metric}, we exemplify that dynamic similarity metric focuses on the exemplar-specific patterns to match similarity. Quantitative results here further demonstrate that the dynamic pattern selection mechanism can improve the naive bilinear similarity metric. 

\begin{table}
  \centering
  \renewcommand{\arraystretch}{1.15} 
  \addtolength{\tabcolsep}{1pt}
    \begin{tabular}{@{}lcccccc@{}}
    \toprule
    No. & SL & SS & SE & DSM & Val MAE & Val MSE  \\ 
    \midrule
    B1 & $\times$ & $\times$   & $\times$ & $\times$  & 19.06  & 67.95 \\
    B2 & \checkmark & $\times$   & $\times$   & $\times$    & 18.32    & 64.01    \\
    B3 &  \checkmark &  \checkmark   & $\times$   &  $\times$   & 17.44    & 67.07    \\
    B4 &  \checkmark & \checkmark &   \checkmark  & $\times$    & 17.35    & 60.28    \\
    B5 & \checkmark &  \checkmark & \checkmark  & \checkmark  & \textbf{15.74}    & \textbf{58.53}    \\ \hline
    
    No. & SL & SS & SE & DSM & Test MAE & Test MSE \\ \hline
    B6 &  $\times$ & $\times$   & $\times$ & $\times$  & 16.71    & 103.31 \\
    B7 & \checkmark  & $\times$   & $\times$   &$\times$     & 14.97    & 92.88    \\
    B8 & \checkmark & \checkmark  & $\times$   & $\times$    & 16.53    & 103.69   \\
    B9 & \checkmark & \checkmark &  \checkmark   & $\times$   & 16.48    & 96.85    \\
    B10 & \checkmark & \checkmark & \checkmark   &  \checkmark    & \textbf{14.62}    & \textbf{91.83}    \\
    \bottomrule
    \end{tabular}
    \vspace{-10pt}
    \caption{\textbf{Ablation study} on self-similarity (SS), dynamic similarity metric (DSM),  similarity loss (SL), and scale embedding (SE).}
    \label{tab:comparison-feature-refiner}
    \vspace{-10pt}
\end{table}

\subsection{Number of Exemplars per Task}
Here we investigate the impact of the number $n$ of exemplars (randomly chosen) per task. Since the given maximum number of exemplars per query image is $3$ in FSC147, we experiment with $n=1,2,3$  and report their results in Table~\ref{tab:different-exemplar-number}. It is foreseeable that more exemplars yield better results as in Table~\ref{tab:different-exemplar-number}. Note that even our method with one single exemplar surpasses the other methods with three exemplars (cf. Table~\ref{tab:comparison-CAC-methods}). This indicates that our method seeing only one exemplar could still capture information to describe the corresponding category.
Besides, CAC methods may get more vulnerable to intra-class variations with fewer exemplars, but we find that self-similarity module offers obvious improvement within our method in this scenario. Refer to supplementary materials for more details.

\begin{table}
  \centering
  \renewcommand{\arraystretch}{1.2}
  \addtolength{\tabcolsep}{2.5pt}
    \begin{tabular}{@{}lccccc@{}}
    \toprule
    $n$ & Val MAE & Val MSE & Test MAE & Test MSE \\ 
    \midrule
    1   & 17.89   & 61.12   & 16.89    &  96.65   \\
    2   & 16.03   & 58.65   & 16.16    &  97.18    \\
    3   & \textbf{15.74}    & \textbf{58.53} & \textbf{14.62} & \textbf{91.83} \\
    \bottomrule
    \end{tabular}
    \vspace{-10pt}
    \caption{\textbf{Impact of the number of exemplars.}}
    \label{tab:different-exemplar-number}
\end{table}

\subsection{How to Integrate Features for the Counter?}
\label{sec:feature-combination}
Here we discuss possible ways to integrate the features before feeding them to the counter. Given the exemplar feature $\boldsymbol z$, the query feature $\boldsymbol x$, and the similarity map $S$, we investigate $4$ ways of feature combination as in Table~\ref{tab:comparison-feature-combination}, where ``+'' denotes channel-wise concatenation. According to the results, only using the similarity map to count objects yields the worst performance (the 1st row) , while leveraging raw features of exemplars and query images can improve the counting performance (the 3rd and 4th rows). However, excluding the similarity map makes the supervision on the similarity metric impossible (the 2nd row). In addition, concatenating the features of exemplars brings marginal improvements but with increased computation overheads (the 3rd row). Therefore, to leverage the information within similarity map while also maintaining a moderate computational cost, we suggest the combination of similarity map and query features as the default representation. 
In supplementary materials, we also show that the query features may encode generic semantic information to help correct the mistakes within the similarity map.

\begin{table}
  \centering
  \renewcommand{\arraystretch}{1.2}
  \addtolength{\tabcolsep}{0.25pt}
  \begin{tabular}{@{}lcccc@{}}
    \toprule
    \small{Combination}  & \small{Val MAE} & \small{Val MSE} & \small{Test MAE} & \small{Test MSE} \\ 
    \midrule
    \scriptsize{${S}$}                           & 21.36   & 69.05   & 18.76 & 92.44     \\
    \scriptsize{$\boldsymbol{x}+\boldsymbol{z}$}        & 19.27   & 66.75   & 18.24 & \textbf{84.39}     \\
    \scriptsize{$\boldsymbol{x}+\boldsymbol{z}+S$}  & \textbf{18.71}   & \textbf{61.88}   & 18.71    & 88.23  \\
    \scriptsize{$\boldsymbol{x}+S$} (default)          & 19.06   & 67.95   & \textbf{16.53} & 103.31  \\
    \bottomrule
    \end{tabular}
    \vspace{-10pt}
    \caption{\textbf{Ways to integrate features for the counter.} $\boldsymbol{x}$ and $\boldsymbol{z}$ stand for the features of query and exemplar, respectively, $S$ for similarity map, and ``+'' for channel-wise concatenation.}
    \label{tab:comparison-feature-combination}
    \vspace{-5pt}
\end{table}

\subsection{Cross-Dataset Generalization}
Following FamNet~\cite{famnet}, we test our models's generality on a car counting dataset CARPK~\cite{carpk}. CARPK contains $1,448$ images of parking lots in a bird view, which differs significantly from the images in FSC147. We exclude the ``car'' category within FSC147 to ensure that training and test categories have no overlap. 

The results are reported in Table~\ref{tab:comparison-carpk}. We first focus on the models without fine-tuning on the CARPK dataset. It can be observed that our models exhibit strong generality. Compared with FamNet, BMNet and BMNet+ obtain a relative performance gain of $49.3\%$ and $63.8\%$ on MAE, respectively. Moreover, BMNet and BMNet+ still retain their advantages when compared with FamNet in the fine-tuning scenario, which demonstrates that our designs are orthogonal to fine-tuning. In addition, the improvements of FamNet and BMNet after fine-tuning indicate the benefit of introducing task-specific information. 

\begin{table}
  \centering
  \renewcommand{\arraystretch}{1.2}
  \addtolength{\tabcolsep}{13.25pt}
    \begin{tabular}{@{}lccc@{}}
    \toprule
    Method    & fine-tuned & MAE   & MSE   \\ 
    \midrule
    FamNet    & \checkmark & 18.19 & 33.66 \\
    BMNet     & \checkmark & 8.05 & 9.70 \\
    BMNet+    & \checkmark & \textbf{5.76}  & \textbf{7.83}  \\ 
    FamNet    & $\times$ & 28.84 & 44.47 \\
    BMNet     & $\times$ & 14.61 & 24.60 \\
    BMNet+    & $\times$ & \textbf{10.44} & \textbf{13.77} \\ \bottomrule
    \end{tabular}
    \vspace{-10pt}
    \caption{\textbf{Generalization performance on the CARPK dataset.} All models are pretrained on the FSC147 dataset. ``fine-tuned'' denotes whether the pretrained models are further fine-tuned on the CARPK dataset.}
    \label{tab:comparison-carpk}
    \vspace{-5pt}
\end{table}

\section{Conclusions and Limitations}
In this work, we show that similarity modeling matters for CAC. In particular, we propose a similarity-aware framework for CAC where the feature representation and similarity metric are jointly learned in an end-to-end manner. Then we instantiate our framework with a naive BMNet that learns bilinear similarity. We also show how to extend the BMNet with the idea of exploiting self-similarity among features, learning dynamic similarity metric, and imposing explicit supervision on the similarity map. Both our BMNet and the extended BMNet+ achieve state-of-the-art performance on the large-scale dataset FSC147 and car counting dataset CARPK.

\margin
\noindent \textbf{Limitations.}
Technically, we mainly focus on designing a better similarity metric, while how to obtain better feature representation is not well addressed: 1) the function of self-similarity module is intuitive, and Table~\ref{tab:comparison-feature-refiner} shows the self-similarity may hurt the performance on the test set; 2) how to integrate rich representation along with similarity map is also not addressed well in this work. Maybe transformer-based  tracking~\cite{transformer_tracking} can help. In addition, since our goal is to present a generic framework, some designs in our instantiated models include some heuristics, which could be further studied in detail. 

\margin
\noindent \textbf{Acknowledgements.} This work was funded by the National Natural Science Foundation of China under Grant No.~61876211 and No.~62106080, and the Chinese Fundamental Research Funds for the Central Universities 
under Grant No.~2021XXJS095.


{\small
\bibliographystyle{ieee_fullname}
\bibliography{egbib}
}

\end{document}